\documentclass[runningheads]{llncs}

\usepackage{eccv}

\usepackage{eccvabbrv}

\usepackage{svg}
\usepackage{graphicx}
\usepackage{booktabs}

\usepackage[accsupp]{axessibility}  

\usepackage{hyperref}

\usepackage{orcidlink}

\usepackage[flushleft]{threeparttable}
\usepackage{multirow}
\usepackage{colortbl}
\usepackage{float}
\begin{document}

\title{Causal Transformer for Fusion and Pose Estimation in Deep Visual Inertial Odometry} 

\titlerunning{Causal Transformer for Fusion and Pose Estimation in Deep VIO}

\author{Yunus Bilge Kurt\inst{1,2,3}\orcidlink{0000-0002-1564-3450} \and
Ahmet Akman \inst{1,2}\orcidlink{0000-0001-5112-6963} \and
A. Aydın Alatan\inst{1,2}\orcidlink{0000-0001-5556-7301
}}

\authorrunning{Y. Kurt et al.}

\institute{Dept. Electrical and Electronics Eng., METU, Ankara, Türkiye \and
Center for Image Analysis (OGAM), METU, Ankara, Türkiye \and 
Codeway AI Research}

\maketitle

\begin{abstract}
In recent years, transformer-based architectures become the de facto standard for sequence modeling in deep learning frameworks.
Inspired by the successful examples, we propose a causal visual-inertial fusion transformer (VIFT) for pose estimation in deep visual-inertial odometry.
This study aims to improve pose estimation accuracy by leveraging the attention mechanisms in transformers, which better utilize historical data compared to the recurrent neural network (RNN) based methods seen in recent methods.
Transformers typically require large-scale data for training. To address this issue, we utilize inductive biases for deep VIO networks.
Since latent visual-inertial feature vectors encompass essential information for pose estimation, we employ transformers to refine pose estimates by updating latent vectors temporally.  
Our study also examines the impact of data imbalance and rotation learning methods in supervised end-to-end learning of visual inertial odometry by utilizing specialized gradients in backpropagation for the elements of SE$(3)$ group.
The proposed method is end-to-end trainable and requires only a monocular camera and IMU during inference. 
Experimental results demonstrate that VIFT increases the accuracy of monocular VIO networks, achieving state-of-the-art results when compared to previous methods on the KITTI dataset. The code will be made available at \url{https://github.com/ybkurt/VIFT}.  
\end{abstract}

\section{Introduction}
\label{sec:intro}

Visual Inertial Odometry (VIO) is a fundamental approach for the estimation of the pose of a moving body by visual-inertial sensor fusion. These methods can leverage the complementary nature of visual and inertial data, where the visual data provides 3D information about the scene, and the inertial data offers robust motion cues. Geometry-based VIO methods show promising results \cite{qin2018vins,campos2021orb,van2023eqvio}; however, they often require careful initialization and calibration to perform accurately during operation. In contrast, end-to-end learning approaches in VIO \cite{clark2017vinet,chen2019selective,yang2022efficient} offer the potential to bypass these challenges by directly learning fuse sensory information. Integrating deep learning into VIO systems introduces the possibility of not only improving accuracy but also simplifying deployment, as the models can generalize across different environments and conditions.

While deep learning has shown promise in VIO, there are still significant areas for improvement, particularly in temporal modeling and rotation estimation. Current deep VIO methods often employ recurrent neural network (RNN) based methodologies to model temporal dependencies. 
Nevertheless, the literature has decent alternatives to RNN-based methods for modeling complex temporal dynamics in VIO tasks. Additionally, rotation regression remains a challenge, as traditional representations such as quaternions or Euler angles are not optimal for deep learning models \cite{zhou2019continuity}. Transformer architecture \cite{vaswani2017transformer} offers a promising alternative for better temporal modeling. By leveraging transformers, we hypothesize that we can achieve more accurate and robust pose estimation, particularly by focusing on refining the latent representations of sensor data and improving the rotation regression.

\begin{figure}[t]
    \centering
    \includegraphics[width=1\linewidth]{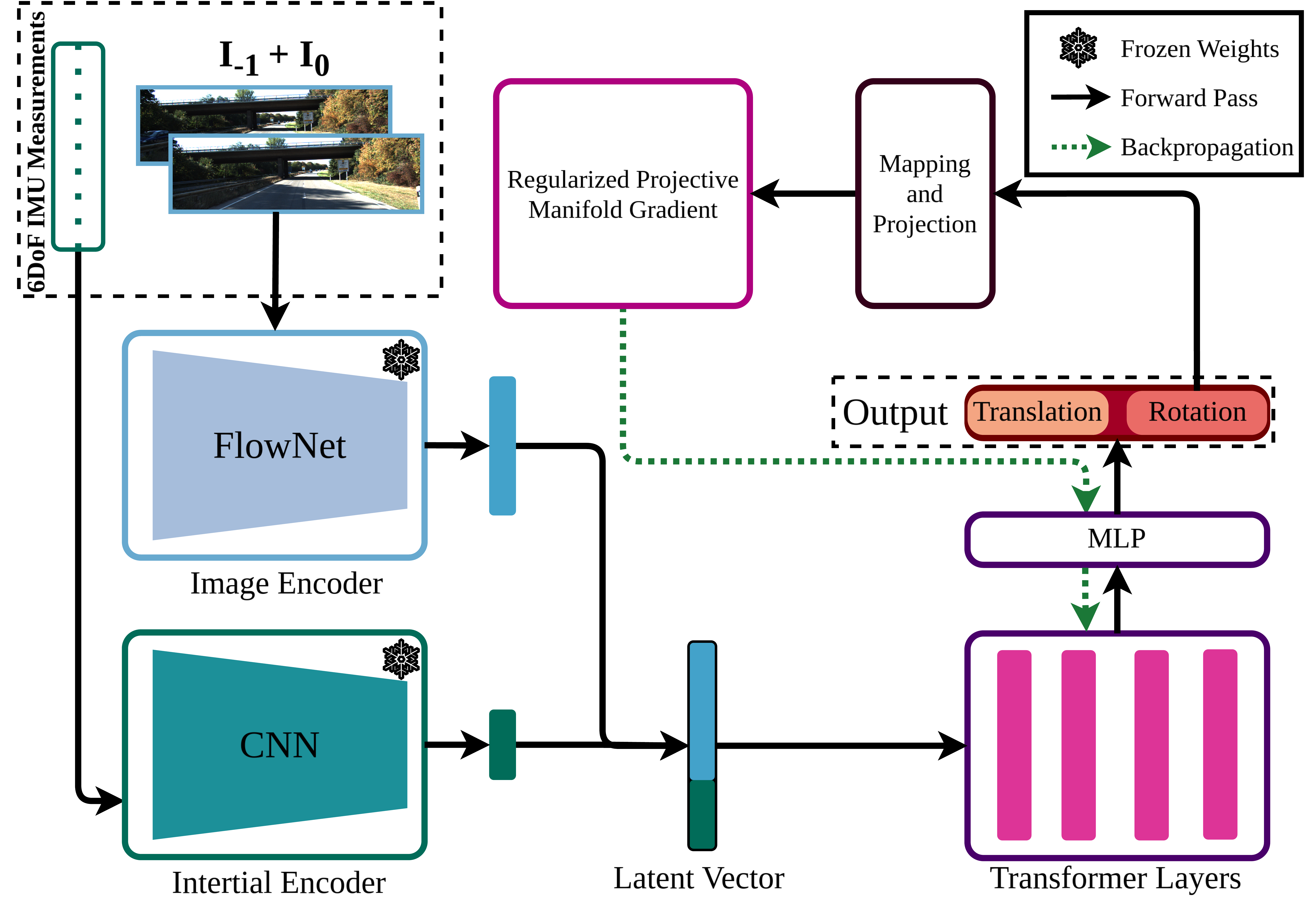}
    \caption{VIFT architecture. The network consists of two fundamental sides. The first side consists of two encoders with frozen weights that map visual and inertial information to a latent space. The second side consists of sequential transformer layers followed by a fully connected layer. For backpropagation enhancement of rotation, the output is projected to $3\times 3$ rotation matrix representation, and RPMG (Regularized Projective Manifold Gradient) \cite{chen2022projective} is used.}
    \label{fig:main-fig}
\end{figure}

As illustrated in Figure \ref{fig:main-fig}, our method uses frozen image and inertial encoders to obtain latent vectors for VIO. Then, transformer layers perform fusion and pose estimation using latent vectors. Our proposed VIFT method is a  ViT-like \cite{Dosovitskiy2020vit} architecture without class token, and instead of image patches, VIFT uses latent visual and inertial representations for learning temporal relations. Instead of creating new vectors from scratch, we modify the latents with the transformer using temporal relations and use output directly to estimate poses. In the last stage, we use RPMG \cite{chen2022projective} for manifold-aware gradient updates for rotation regression. We show that the VIFT improves the performance of deep VIO networks compared to RNN-based pose modules.

We propose visual-inertial fusion transformer (VIFT), a novel fusion and pose estimation module for VIO based on causal transformer encoder architecture. Our contributions can be summarized as follows:
\begin{itemize}
    \item VIFT uses transformer layers for visual-inertial fusion and pose estimation. We find inductive biases to make transformers perform better, eliminating the possible problems arising from the small data scale. 
    
    \item VIFT exploits Riemannian manifold optimization techniques for rotations, enabling the network to learn the rotations better than Euler angles and quaternions used in previous works.
    
    \item VIFT achieves state-of-the-art results with the proposed transformer module and improves performance further with manifold-aware gradients. 
\end{itemize}

\section{Related Work}

\subsection{Visual-Inertial Odometry}
Visual-inertial odometry (VIO) leverages the complementary strengths of visual and inertial sensors to provide robust and accurate motion estimation. While visual odometry relies on camera images to estimate motion, it can fail in textureless environments. In contrast, inertial odometry offers high update rates using data from Inertial Measurement Units (IMUs) but requires proper initialization and is prone to drift due to biases in accelerometer and gyroscope readings. With fusion visual and inertial estimates, VIO can mitigate the weaknesses of each individual sensor. Visual measurements help correct the drift in inertial measurements by providing pose updates, while inertial measurements enhance the robustness and temporal resolution of visual estimates. This has led to the development of several VIO methods.

Geometry-based VIO methods require addressing several challenges, including the excitation of IMU biases along all axes to track biases, proper camera-IMU extrinsic calibration, and robust initialization procedures. Despite these challenges, the fusion of visual and inertial data through sophisticated filtering based \cite{mourikis2007multi,bloesch2015robust,van2023eqvio} and factor graph based \cite{leutenegger2015keyframe,forster2016manifold,campos2021orb,qin2018vins,stumberg22dmvio} methods has significantly advanced the field, enabling more accurate and reliable motion estimation in various applications.

Geometry-based visual inertial odometry methods have several drawbacks. They require good initialization with excitation in all axes to determine IMU biases, which might not be feasible in many scenarios. Additionally, tuning various parameters is necessary for robust and accurate operation. 

VINet \cite{clark2017vinet}, being a seminal work in the field, approached visual-inertial odometry as a sequence-to-sequence learning problem. This led to development of supervised deep VIO methods \cite{clark2017vinet,chen2019selective,liu2021atvio,tu2022ema,yang2022efficient,wang2022attention} which use ground truth transformation to learn correct 
transformation via regression, and self-supervised deep VIO methods \cite{shamwell2019unsupervised,han2019deepvio,wei2021unsupervised,wagstaff2022self,liu2021atvio,almalioglu2022selfvio} which perform view synthesis with estimated relative pose between images and train on pixel-wise intensity errors. 

In supervised deep VIO, Chen et al. \cite{chen2019selective} proposed Soft Fusion and Hard Fusion for adaptively weighting visual and inertial embeddings during inference. Yang et al. \cite{yang2022efficient} proposed training an adaptive modality selection module inside a deep VIO network to optionally disable the visual encoder. Until recent years, fusion and pose estimation modules consisted of RNN-based networks. ATVIO \cite{liu2021atvio} suggested using the attention mechanism for fusion in VIO. External memory-aided attention-based module is proposed in EMA-VIO \cite{tu2022ema}. Recent methods showed improvements in VIO performance with attention-based modules.

The recent advancements induced interest in exploring transformer-based architectures for fusion and pose estimation in VIO. Transformers, known for their ability to model long-range dependencies and capture complex temporal dynamics, present a compelling alternative to traditional RNN-based approaches. By leveraging self-attention mechanisms, transformers can selectively focus on the most relevant parts of the latent visual-inertial vectors, potentially leading to a more accurate and robust fusion of visual and inertial data.

\section{Method}

End-to-end VIO methods consist of a visual encoder, an inertial encoder, a fusion module, and a pose estimation module. The visual encoder extracts visual features from consecutive frames to provide 3D understanding of their architecture. The inertial encoder takes input from the IMU measurements between frames. The rate of IMU data is usually higher than that of cameras.  
The fusion module takes the visual and inertial representation presented by encoders. The pose estimation module uses fused representations of visual and inertial information to estimate the pose.

\subsection{Feature Encoder}

VIFT architecture can be seen in Figure \ref{fig:main-fig}. At each timestep $t$, VIFT takes two consecutive frames $\mathbf{I}_t, \mathbf{I}_{t-1} \in \mathcal{R}^{C\times W \times H}$ and IMU measurements $\mathbf{a}_{t,t-1}, \mathbf{\omega}_{t,t-1}$ between frames, consisting of accelerometer and gyroscope readings.
Visual and inertial measurements are processed with different encoders for their modality as in previous deep VIO methods \cite{chen2019selective, yang2022efficient}, following the work of \cite{yang2022efficient} we use FlowNet \cite{dosovitskiy2015flownet} based image encoder and 1D CNN based inertial encoder.
Visual measurements are processed by image encoder $\mathbf{E}_v$ to produce one-dimensional visual encodings $\mathbf{x}^v_t$.
\begin{equation}
    \mathbf{x}^v_t = \mathbf{E}_v(\mathbf{I}_t, \mathbf{I_{t-1}})
\end{equation}
Inertial measurements are processed by inertial encoder $\mathbf{E}_i$ to produce one-dimensional inertial encodings $\mathbf{x}^i_t$. 
\begin{equation}
    \mathbf{x}^i_t = \mathbf{E}_i(\mathbf{a}_{t,t-1}, \mathbf{\omega}_{t,t-1})
\end{equation}
At the end, we concatenate visual and inertial latent vectors. 
\begin{equation}
    \mathbf{x}_t = \textit{Concat}(\mathbf{x}^v_t, \mathbf{x}^i_t)
\end{equation}
The corresponding $\mathbf{x_t}$ is labeled as Latent Vector in Figure \ref{fig:main-fig}. The concatenated vector is passed to the transformer-based fusion and pose estimation module. The expected output is the translation and rotation of the camera between two timesteps $\mathbf{T}_{t-1}^{t}\in \textbf{SE}(3)$.
 
 \begin{figure}[t]
     \centering
     \includegraphics[width=1\linewidth]{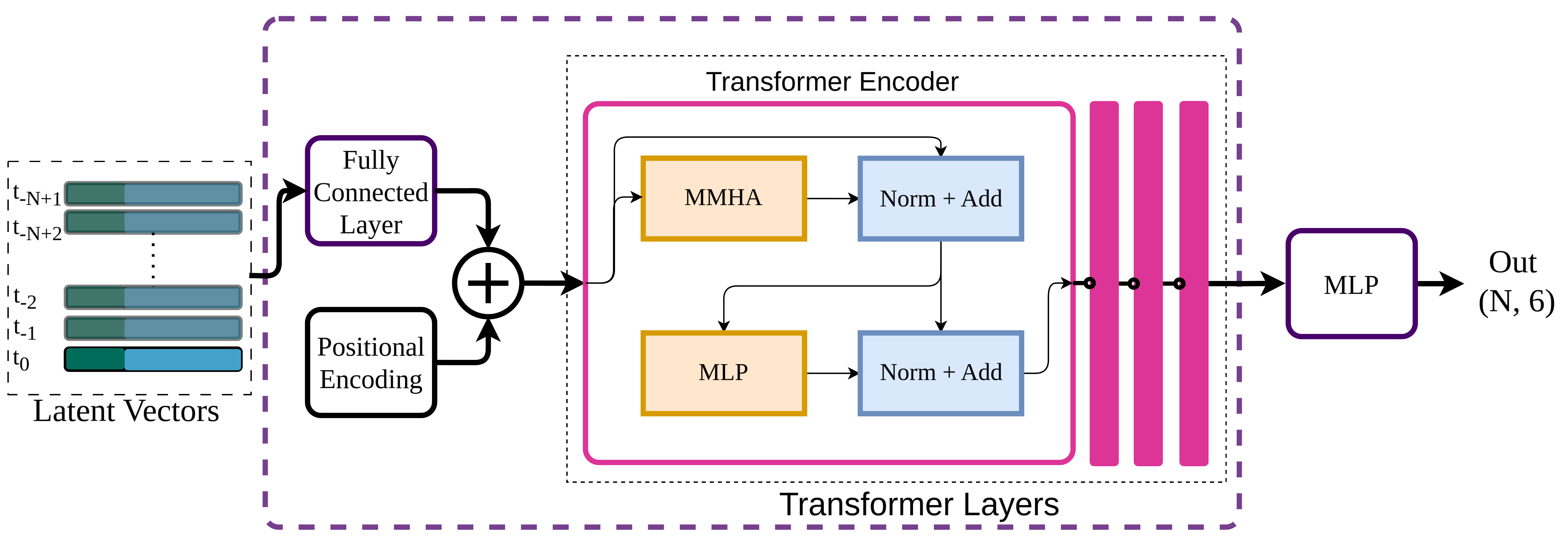}
     \caption{Causal transformer based architecture for fusion and pose estimation.}
     \label{fig:transformers}
 \end{figure}

\subsection{Transformers For Fusion and Pose Estimation}

We use transformer encoder layers with causal masks to update latent vectors for pose estimation. The fusion and pose estimation module modifies each latent vector with an attention mechanism. Weighting the latent vectors based on data-dependent masks is introduced in Soft Fusion \cite{chen2019selective}, where a soft mask function is determined based on the current latent visual inertial vector.
The difference in VIFT is determining the mask based on previous measurements in the local window and applying several masks multiple times through the layers of the transformer. In the end, the corresponding vector is modified with past information to provide more accurate pose estimation. 

The fusion and pose estimation module of VIFT can be seen in Figure \ref{fig:transformers}. In Figure \ref{fig:transformers}, the input latent vectors $\mathbf{x}_{t}, \dots, \mathbf{x}_{t-N}$ are shown on the left. In the first step, similar to ViT, we apply linear projection to visual-inertial latent vector $\mathbf{x}_t$. This projection applies learned weights to visual and inertial encodings. We keep the embedding dimension of the transformer the same with a concatenated visual inertial latent vector. 

After this step, transformer Layers take $\mathbf{x}_t$ together with N-1 past measurements  $\mathbf{x}_{t-1}, \dots, \mathbf{x}_{t-N}$ as input.
Then, we add Sinusoidal Positional Encodings to each latent vector according to their location in the sequence.
No temporal information flows between the linear projection's latent visual inertial vector sequence and the Positional Encoding steps.
Transformer Layers modify the resulting latent vectors with attention by considering a local window.
In this step, latent vectors are weighted based on the previous and current measurements.
The operation of the transformer layer can be summarized as follows.
A transformer layer consists of a masked multi-head attention (MMHA) layer, feed-forward layer, residual connections, and layer normalizations \cite{vaswani2017transformer}.
The attention operation computes a weighted sum of value vectors based on the normalized dot product of query and key vectors. With a causal mask, we ensure that current estimates are not influenced by future measurements, which is crucial for real-time applications.

After transformer layers, we apply 2-layer MLP to every feature to obtain pose output. We follow Yang et al. \cite{yang2022efficient} and give a 6-dimensional output consisting of translation and Euler angles for rotation. Ultimately, we obtain $N$ relative poses for $N+1$ input images. 
During inference, for the first $N+1$ images, we use all the output pose estimates to initialize the process and take only the last estimate after shifting latent vectors and adding a new latent visual inertial vector to the end of the sequence.

\subsection{Deep Rotation Regression}

The manifold structure of \textbf{SO}$(3)$ space should be considered while performing optimizations on rotations. Zhou et al. \cite{zhou2019continuity} discuss the importance of continuous representations in neural network optimization. A continuous subset of rotations can be discontinuous in the Euler angle representation of rotations, which could create discontinuous training signals in training. Moreover, the interpolation problems in Euler angles and gimbal lock phenomena further motivate using the optimization techniques for manifolds.   

We use the Regularized Projective Manifold Gradient (RPMG) layer proposed by Chen et al. \cite{chen2022projective} to obtain training signals for rotations, which produces manifold-aware gradients in backward passes. We first convert our 3D estimation to a 9D rotation matrix. Then, we calculate the loss between ground truth rotation $\mathbf{R}_{\text{gt}}$ and estimated rotation $\mathbf{R}$.
For translations, we use direct regression as the space is already Euclidean. The total loss is the weighted sum of rotation and translation losses.
\begin{equation}
\label{eq:loss}
    \mathcal{L}_r = ||\mathbf{R} - \mathbf{R}_{\text{gt}}||, \quad \mathcal{L}_t = ||\mathbf{t} - \mathbf{t}_{\text{gt}}||, \quad \mathcal{L} = \mathcal{L}_t + \alpha\mathcal{L}_r
\end{equation}

where $\alpha$ is a constant factor to balance rotational and translational loss terms and $|| \cdot ||$ is a norm. In the backward pass, we calculate RPMG \cite{chen2022projective} for rotation loss, which uses Riemannian optimization to get a goal rotation $\mathbf{R}_g$ and maps it back to the representation manifold to find the closest element of ambient space to estimation, which is used for obtaining the gradients in the backward pass.

\section{Experiments}

We evaluate VIFT and our choices in training settings and model selection. We show transformer based modules are effective in fusion and pose estimation in deep VIO.

\subsection{Experiment Setup}

We utilize KITTI Odometry Dataset \cite{kitti} for training and benchmarking in this study. The dataset is widely used amongst the visual-inertial odometry research community and consists of 22 sequences where those sequences have stereo-recorded images and 6-Degree-of-Freedom IMU measurements. Following \cite{chen2019selective, yang2022efficient}, we train our method with Sequences 00, 01, 02, 04, 06, and 09 while choosing Sequences 05, 07, and 10 for testing. We do not use Sequence 03 as it misses the IMU. The input images and ground truth poses are recorded at a rate of 10 Hz, whereas IMU data is recorded at 100 Hz. As the challenge requires monocular images, only left-camera frames are used throughout the study. For evaluation metric, relative translation error $\textit{t}_{\textit{rel}}$ and relative rotation error $\textit{r}_{\textit{rel}}$ are calculated, indicating the averaged translation and rotation drift of all subsequences with length of (100 m, ..., 800 m).

We utilize pretrained FlowNet-S \cite{dosovitskiy2015flownet} based image encoder and 1D CNN based inertial encoder from Yang et al. \cite{yang2022efficient}, and keep them fixed during training. Input images are resized to $512\times256$ resolution. Our training scheduler follows a cosine annealing learning rate with warm restarts \cite{loshchilovSGDR}, with restarts occurring every 25 epochs. We employ the AdamW \cite{adamw} optimizer with a learning rate of \(1 \times 10^{-4}\), we set $\beta_{1}=0.9$, $\beta_{2}=0.999$. The loss function in Equation \ref{eq:loss} includes a rotation weight of \(\alpha = 40\) and uses the L1 norm. We use RPMG layer for rotation estimation with \(\tau = \frac{1}{4}\) and \(\lambda = 0.01\). We use a sequence length of $N=11$ during training.

Our transformer model is configured with an embedding dimension of 768 and a feed-forward layer dimension of 128. The embedding dimension is selected to be equal to the sum of 512 and 256, which are the output dimensions of the image encoder and inertial encoder, respectively. The model uses Sinusoidal Positional Encodings, consists of 4 transformer encoder layers with 6 attention heads, and employs masked self-attention with a causal mask. We do not apply dropout in the transformer. The network is trained for 200 epochs with a batch size of 128, totaling 27k training steps. Experiments are conducted using an NVIDIA GeForce RTX 4060 Laptop GPU.

\subsection{Main Results}
\begin{table}[t]

\centering
\caption{Comparison with prior VIO works in translational \& rotational error metrics of KITTI Odometry Benchmark \cite{kitti}. The best performances in each block are marked in \textbf{bold} and overall bests are shown with \colorbox{green!25}{green} background. Loop closure is excluded for VINS-Mono. Results are taken from Yang et al. \cite{yang2022efficient} except ours.}

\label{table:main_results}
\begin{threeparttable}
\begin{tabular}{cc|cc|cc|cc}
\hline
& \multirow{2}{*}{Method}        & \multicolumn{2}{c|}{Seq. 05} & \multicolumn{2}{c|}{Seq. 07} & \multicolumn{2}{c}{Seq. 10}  \\
           &  & {\tiny $t_{rel}(\%)$}       & {\tiny$r_{rel}(^{\circ})$}      & {\tiny $t_{rel}(\%)$}     & {\tiny$r_{rel}(^{\circ})$}      & {\tiny $t_{rel}(\%)$}        & {\tiny$r_{rel}(^{\circ})$}   \\ \hline\hline
Geo & VINS-Mono \cite{qin2018vins}       & \textbf{11.6 }     & \textbf{1.26 }      & \textbf{10.0}       & \textbf{1.72 }     & \textbf{16.5}       & \textbf{2.34}    \\ \hline
\multirow{2}{*}{\begin{tabular}{@{}c@{}c@{}} Self- \\ Sup. \end{tabular}} &VIOLearner \cite{shamwell2018vision}      & 3.00      & \textbf{1.40}        & 3.60      & 2.06       & 2.04       & 1.37    \\
& DeepVIO \cite{han2019deepvio}       & \textbf{2.86 }     & 2.32       & \textbf{2.71}       & \textbf{1.66}       & \cellcolor{green!25}\textbf{0.85}       & \textbf{1.03}      \\ \hline
\multirow{6}{*}{Sup.} & ATVIO \cite{liu2021atvio}      & 4.93      & 2.4       & 3.78       & 2.59       & 5.71       & 2.96       \\
&Soft Fusion \cite{chen2019selective}      & 4.44      & 1.69       & 2.95       & 1.32      & 3.41       & 1.41        \\
&Hard Fusion \cite{chen2019selective}      & 4.11      & 1.49       & 3.44       & 1.86       & \textbf{1.51}       & 0.91       \\
&Yang et al. \cite{yang2022efficient}          & 2.01       & 0.75       & 1.79       & 0.76       & 3.41       & 1.08        \\ \cline{2-8}
&(Ours) Baseline \;       & \cellcolor{green!25}\textbf{1.93}       & 0.68       & \cellcolor{green!25}\textbf{1.55}      & 0.91       & 2.57       & 0.54         \\ 
&(Ours) w. RPMG   \;       & 2.02       & \cellcolor{green!25}\textbf{0.53 }      & 1.75       & \cellcolor{green!25}\textbf{0.47}       & 2.11       & \cellcolor{green!25}\textbf{0.39 }        \\ \hline 
\end{tabular}
\end{threeparttable}
\end{table}

Table \ref{table:main_results} compares our method against geometry-based, self-supervised, and supervised VIO approaches. Self-supervised methods \cite{shamwell2018vision,almalioglu2022selfvio} are trained on KITTI sequences $00$ to $08$. We include supervised methods that utilize the same training and testing splits as our approach \cite{liu2021atvio, chen2019selective,yang2022efficient} on the KITTI dataset \cite{kitti}.

Monocular VIO with geometry-based methods requires excitation of all axes in initialization to correctly determine IMU biases and scale. Cars in the KITTI dataset mostly move forward and rotate in the yaw axis. This type of motion makes it hard to evaluate their performance fairly for VIO methods that require IMU initialization. ORB-SLAM3 \cite{campos2021orb} does not initialize in monocular inertial mode, and VINS-Mono \cite{qin2018vins} produces high errors even if it can initialize.

As seen in Table \ref{table:main_results}, VIFT obtains state-of-the-art performance compared to learning-based methods. VIFT, without other additional modules, provides the lowest translation errors in Sequences $05$ and $07$ and the lowest rotation errors in Sequences $05$ and $10$ while obtaining comparable performances in other metrics. Moreover, with RPMG \cite{chen2022projective}, VIFT decreases the rotation error by $\approx63.8\%$ in test Sequence 10 compared to Yang et al. \cite{yang2022efficient}. Our experiments show transformer-based fusion and pose estimation surpass the performance of methods that use the same visual and inertial features with the RNN-based networks.

\subsection{Ablation Study}

\begin{figure}[htb]
 \centering
 \includegraphics[width=1\linewidth]{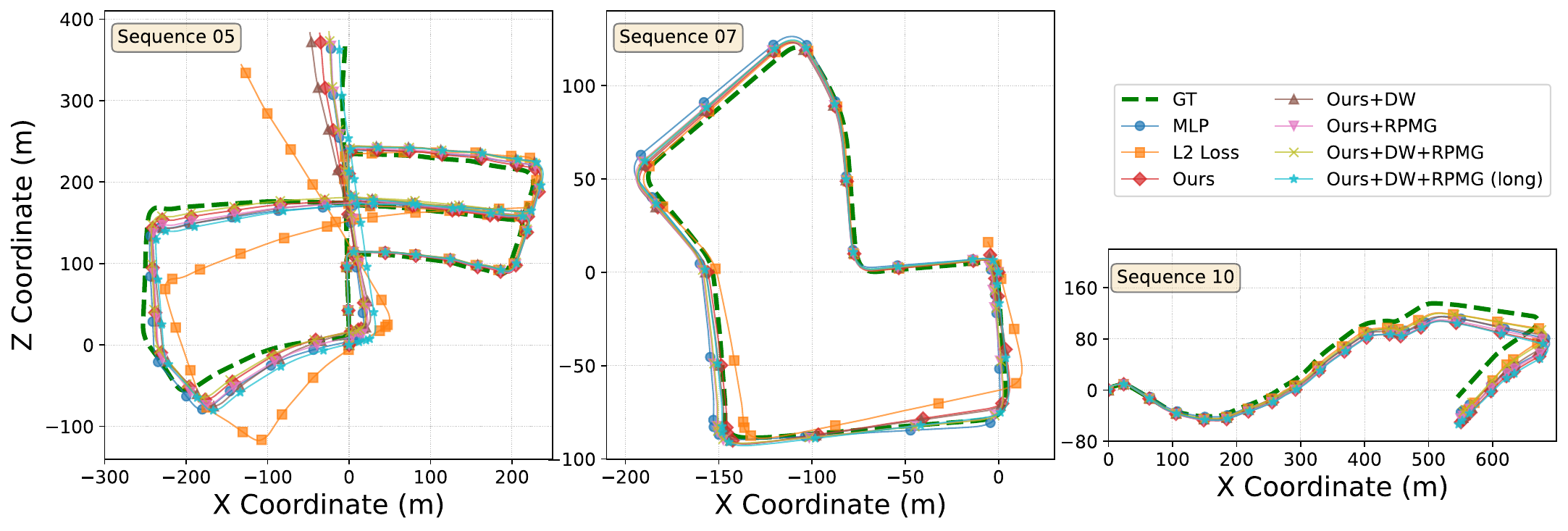}
 \includegraphics[width=1\linewidth]{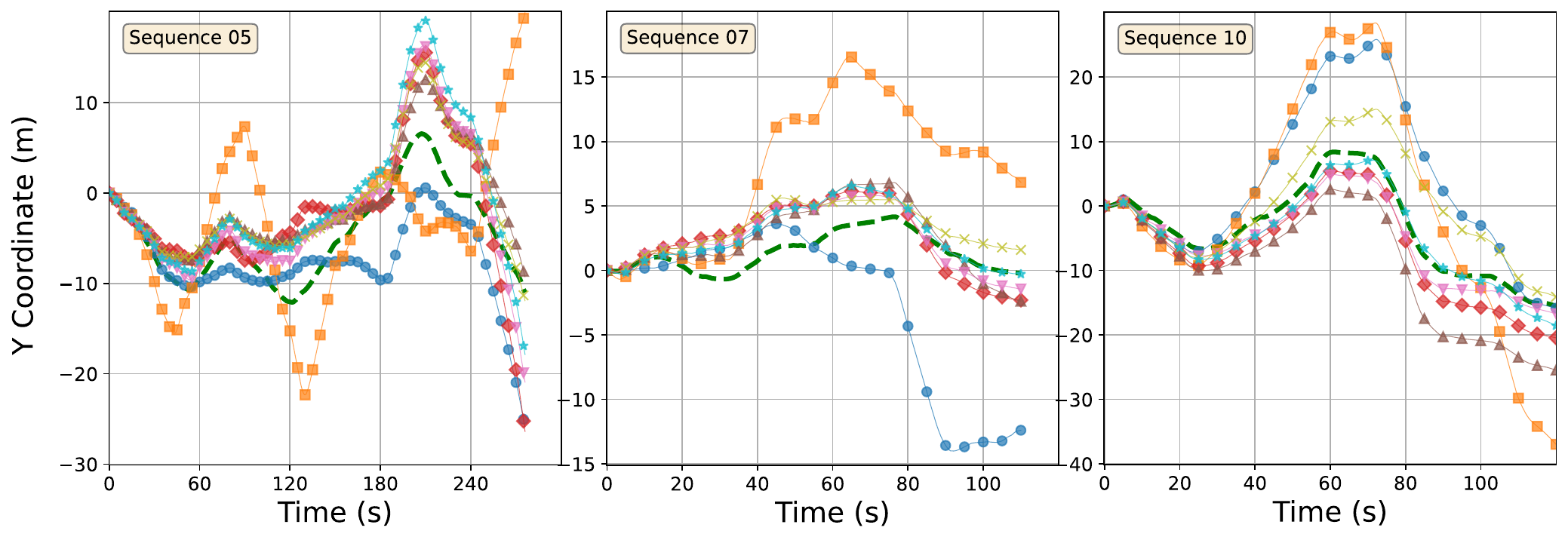}
 \caption{Proposed transformer based fusion and pose estimation module in VIFT evaluated under different training settings. We mark trajectories every 5 seconds for intuition about the vehicle's speed along the trajectory and easy distinction of results. We emphasize that the camera and IMU provide 10 FPS and 100 Hz measurements, respectively, which are much more frequent than marked locations. We show the estimated trajectory in test sequences from above in the top row and vertical trajectory versus the bottom row. All trajectories start from the origin, and relative pose estimates from VIFT are applied sequentially to obtain absolute pose estimates for each time index.} 
 \label{fig:results}
\end{figure}

In this section, we look at the effect of different modules to understand the performance of VIFT. We show the KITTI evaluation metric results in Table \ref{tab:ablation_study} and plot the estimated trajectories against ground truth trajectories in Figure \ref{fig:results}. We mark trajectories every 5 seconds to obtain intuition about the vehicle's speed along the trajectory and to make it easy to distinguish results. We emphasize that the camera and IMU provide 10 FPS and 100 Hz measurements, respectively, which are much more frequent than marked locations. We show the estimated trajectory in test sequences from above in the top row and vertical trajectory versus the bottom row. All trajectories start from the origin, and relative pose estimates from VIFT are applied sequentially to obtain absolute pose estimates for each time index. 

\begin{table}[t]
  \caption{Ablation study. Modified parts from VIFT model are shown with \colorbox{green!25}{green} background. Best results are shown \textbf{bold}.}
  \label{tab:ablation_study}
  \centering
  \small
  \begin{tabular}{c|c|c|c|c|cc|cc|cc}
    \hline
     Model & Sequence  & \multirow{2}{*}{Criterion} & Data & \multirow{2}{*}{RPMG} & \multicolumn{2}{c|}{Seq. 05} & \multicolumn{2}{c|}{Seq. 07} & \multicolumn{2}{c}{Seq. 10} \\
    Type & Length & & Balancing & & {\tiny $t_{rel}(\%)$} & {\tiny$r_{rel}(^{\circ})$} & {\tiny $t_{rel}(\%)$} & {\tiny$r_{rel}(^{\circ})$} & {\tiny $t_{rel}(\%)$} & {\tiny$r_{rel}(^{\circ})$} \\ \hline\hline
    \multirow{1}{*}{\cellcolor{green!25}MLP}  
    & \cellcolor{green!25}2  & L1 & & \cellcolor{green!25} & 2.03 & 0.67 & 3.04 & 1.19 & 3.60 & 1.14 \\
    \hline
    \multirow{6}{*}{Ours}   
    & 11 & \cellcolor{green!25}L2 &  & \cellcolor{green!25} & 4.35 & 1.91 & 2.97 & 2.23 & 3.66 & 1.92 \\
    & 11 & L1 &  & \cellcolor{green!25} & 1.93 & 0.68 & \textbf{1.55} & 0.91 & 2.57 & 0.54 \\
    & 11 & L1 & \cellcolor{green!25}\checkmark & \cellcolor{green!25} & 2.37 & \textbf{0.52} & \textbf{1.55} & 0.85 & 2.32 & 0.75 \\
    & \cellcolor{green!25}65 & L1 & \cellcolor{green!25}\checkmark & \checkmark & 2.37 & 0.64 & 1.98 & 0.58 & 2.97 & 0.69 \\
    & 11 & L1 & \cellcolor{green!25}\checkmark & \checkmark & \textbf{1.90} & 0.53 & 1.79 & \textbf{0.45} & 2.40 & 0.60 \\
    & 11 & L1 &  & \checkmark & 2.02 & 0.53 & 1.75 & 0.47 & \textbf{2.11} & \textbf{0.39} \\
    
   \hline
  \end{tabular}
\end{table}

\subsubsection{Model Type}
We first look at the performance of 4-layer MLP trained on latent visual inertial feature vectors. From results in rows 1 and 2 of Table \ref{tab:ablation_study}, we observe that the odometry performance is reasonably good even with a small MLP network. This performance supports the primary motivation of our architecture. The vectors in latent space already contain good properties for pose estimation. We use transformer-based fusion to correct these latent vectors with the transformer, based on past measurements, and a 2-layer MLP is used at the end of the transformer. VIFT utilizes history to improve pose estimation with transformer-based architecture.

\subsubsection{Norm Type in Training Criterion} We found that using the L1 loss function resulted in better performance within the same training steps. As the errors decrease after the initial epochs, the gradients in L2 loss become smaller, leading to slower convergence and requiring more training iterations, according to our observations. Consequently, we observed that training with L2 loss was slower overall. We also tuned each scenario's $\alpha$ parameter in the Equation \ref{eq:loss}. Following previous work \cite{chen2019selective, yang2022efficient}, we used $\alpha=100$ with the L2 criterion. For L1 loss, we found that $\alpha=10$ worked better for Euler angles, while $\alpha=40$ yielded better results in models incorporating the RPMG layer. We experimented with different $\alpha$ values for each method and reported the results using the values that best balanced rotational and translational errors. Since the rotation loss is calculated based on the mean difference between elements of rotation matrices in RPMG \cite{chen2022projective}, we fine-tuned the $\alpha$ parameter to identify the optimal balance.

\subsubsection{Data Balancing}

In datasets like KITTI \cite{kitti}, specific rotational movements, such as sharp turns and sudden stops, are underrepresented. These motions are critical because they impact overall trajectory accuracy more, making errors in these scenarios more costly. To address this, we experimented with increasing the weights of these less frequent rotational updates during training with the histogram of rotations proposed by Yang et al. \cite{yang2021delving}, aiming to enhance the model's performance in these critical cases.

A comparison of rows 3,4,6, and 7 in Table \ref{tab:ablation_study} shows that introducing data balancing during training does not lead to consistent improvements across the test sequences. These results indicate that data balancing might negatively affect the model's ability. Although we obtained the best results in Sequences 05 and 07 when we applied data balancing, the improvements are inconsistent across all sequences.

Our observations suggest that while data balancing can be a helpful strategy, it must be carefully tuned, especially when combined with advanced optimization techniques like RPMG. Overemphasizing underrepresented rotations does not always yield the desired improvements and could potentially degrade performance, particularly in orientation accuracy. Therefore, a more nuanced approach may be required to balance the representation of various movements in the training data without compromising the model's overall robustness.

\subsubsection{RPMG}

The RPMG \cite{chen2022projective} layer has proven to be a crucial enhancement in the models we evaluated, particularly in terms of reducing orientation errors, denoted as $\textit{r}_{\textit{rel}}$. When comparing models trained with RPMG to those without, we consistently observe significant improvements in orientation accuracy across all tested sequences. This consistent reduction in $\textit{r}_{\textit{rel}}$ highlights the effectiveness of incorporating Riemannian optimization techniques in the training process.

RPMG is particularly advantageous in scenarios involving rotations, where traditional optimization methods might struggle due to the non-Euclidean nature of orientation spaces. By operating directly on the manifold of rotations, RPMG ensures that updates to the orientation parameters are more geometrically appropriate, leading to better convergence properties and, ultimately, more accurate predictions.

In our ablation study, including RPMG improved orientation accuracy and demonstrated robustness across different sequence lengths and data balancing strategies. For instance, in Sequence 07, the use of RPMG led to a marked decrease in $\textit{r}_{\textit{rel}}$, from 0.91 to 0.47, highlighting its substantial impact. We observed similar trends across Sequence 05 and Sequence 10, where RPMG consistently yielded lower orientation errors.

\subsubsection{Sequence Length}  When we compare models in rows 5 and 6 of Table \ref{tab:ablation_study}, we observe a performance drop on the same model trained with increased sequence length. We tried a sequence length of 65 compared to the original sequence length of 11 in the experiment. Modeling the relationships in longer sequences is a more complex task, and larger sequence lengths could require more training data.

\section{Conclusion}

We introduce deep VIO network VIFT, which performs sensor fusion and pose estimation with a causal transformer. We show that our method outperforms previous methods with our experiments on the KITTI dataset. We also improve the VIFT by including the manifold optimization technique RPMG inside our pipeline. Our study's consistent performance gains across different configurations demonstrate that VIFT and RPMG provide a fundamental enhancement that can significantly elevate the performance of visual-inertial odometry models.

\section*{Acknowledgements}
We gratefully acknowledge the computational resources provided by TÜBİTAK ULAKBİM High Performance and Grid Computing Center (TRUBA). Yunus Bilge Kurt is supported by the TÜBİTAK under the 2210-National MSc/MA Scholarship Program.  

%
%
\bibliographystyle{splncs04}
\bibliography{main}
\end{document}